\documentclass[runningheads]{llncs}

\usepackage[final,year=2026]{accv}

\usepackage{accvabbrv}
\usepackage{graphicx}
\usepackage{booktabs}
\usepackage{array}
\usepackage{placeins}
\usepackage{iftex}
\ifPDFTeX
  \usepackage[accsupp]{axessibility}
\fi
\usepackage[hidelinks]{hyperref}
\usepackage{orcidlink}

\newcommand{\gain}[1]{+#1}
\newcommand{\best}[1]{\textbf{#1}}
\begin{document}

\title{Adaptive Observation Control for Active Visual Recognition}
\titlerunning{Adaptive Observation Control}

% Shared publication metadata for the main paper and supplement.
% The shared star identifies Weijia Wang and Yuan-Gen Wang as corresponding authors.
% Dikai Li's real email is pending; the visible placeholder is intentional.
\author{%
Dikai Li \and
Weijia Wang\textsuperscript{*} \and
Yuan-Gen Wang\thanks{Corresponding authors: Weijia Wang and Yuan-Gen Wang.
This work was partially supported by the NSFC 62502108.}%
}

\authorrunning{D.~Li et al.}

\institute{%
School of Artificial Intelligence, Guangzhou University, Guangzhou, China\\
\email{dikaili@e.gzhu.edu.cn, wwj@gzhu.edu.cn, wangyg@gzhu.edu.cn}%
}

\hypersetup{pdfauthor={Dikai Li, Weijia Wang, Yuan-Gen Wang}}

%\inst{1}

\maketitle

\begingroup
\renewcommand{\thefootnote}{}
\footnotetext[0]{This paper has been accepted for presentation in ACCV 2026.}
\endgroup

\begin{abstract}
Visual recognition models usually process a fixed view under a fixed computation budget, although images often benefit from different scales and amounts of spatial evidence. Active vision can adapt by observing a coarse scene and then selecting informative fixations, but the public AdaptiveNN-DeiT-S pipeline still fixes both field of view and observation count. We extend this fixed-step recognizer with a \emph{controllable observation interface} that adapts both the field of view and the number of fixations to the input. Specifically, the model predicts a per-fixation scale that sets its field of view, refines each selected crop with optional scale-gated foveation and a residual high-resolution detail-token branch, and uses label-grounded stop heads for input-dependent early termination. Experiments across fine-grained benchmarks, full ImageNet-1K, and ImageNet-A/R-derived classification splits show that the interface remains competitive with fixed-step references while exposing a controllable observation budget. The same checkpoint enables threshold-controlled accuracy-compute trade-offs without per-threshold retraining. On sequential ImageNet-1K, the learned stop rule outperforms fixed-step truncation of the same checkpoint, sustaining 81.9\% top-1 using 2.45 of the four focus steps and exceeding three-step truncation at lower cost. On each benchmark, a variant combining scale control and detail-token fusion achieves the highest accuracy among the tested component-added non-exit variants. Our code is available at https://github.com/GZHU-DVL/LI-Dikai.
\end{abstract}

\section{Introduction}
\label{sec:intro}

Modern image classifiers typically process a fixed-resolution view of the whole image. This design is efficient on parallel hardware, but it assigns the same observation budget to easy and hard examples, and small discriminative evidence can be diluted by background clutter. Active visual recognition offers a complementary design, in which the model first builds a global representation, then samples task-relevant regions before deciding. AdaptiveNN~\cite{wang2025adaptivenn} instantiates this coarse-to-fine idea, yet the public DeiT-S implementation still uses a fixed crop size across a fixed number of focus steps.

This fixed schedule imposes three limitations: (i)~the field of view is fixed, although discriminative evidence ranges from local texture to larger object parts; (ii)~every crop is encoded uniformly, so high-resolution local detail can be missed and peripheral context can be overemphasized; and (iii)~the observation count is fixed, so easy examples still incur the full computational cost of the focus sequence. We address these limitations by adding a controllable observation interface to the same recognizer while maintaining competitive recognition accuracy. We organize the interface around the corresponding three observation decisions: \emph{geometry control} determines the field of view of each fixation, \emph{evidence control} determines how selected regions are encoded, and \emph{stopping control} decides how many observations an input needs.

\begin{figure}[t]
  \centering
  \includegraphics[width=\linewidth]{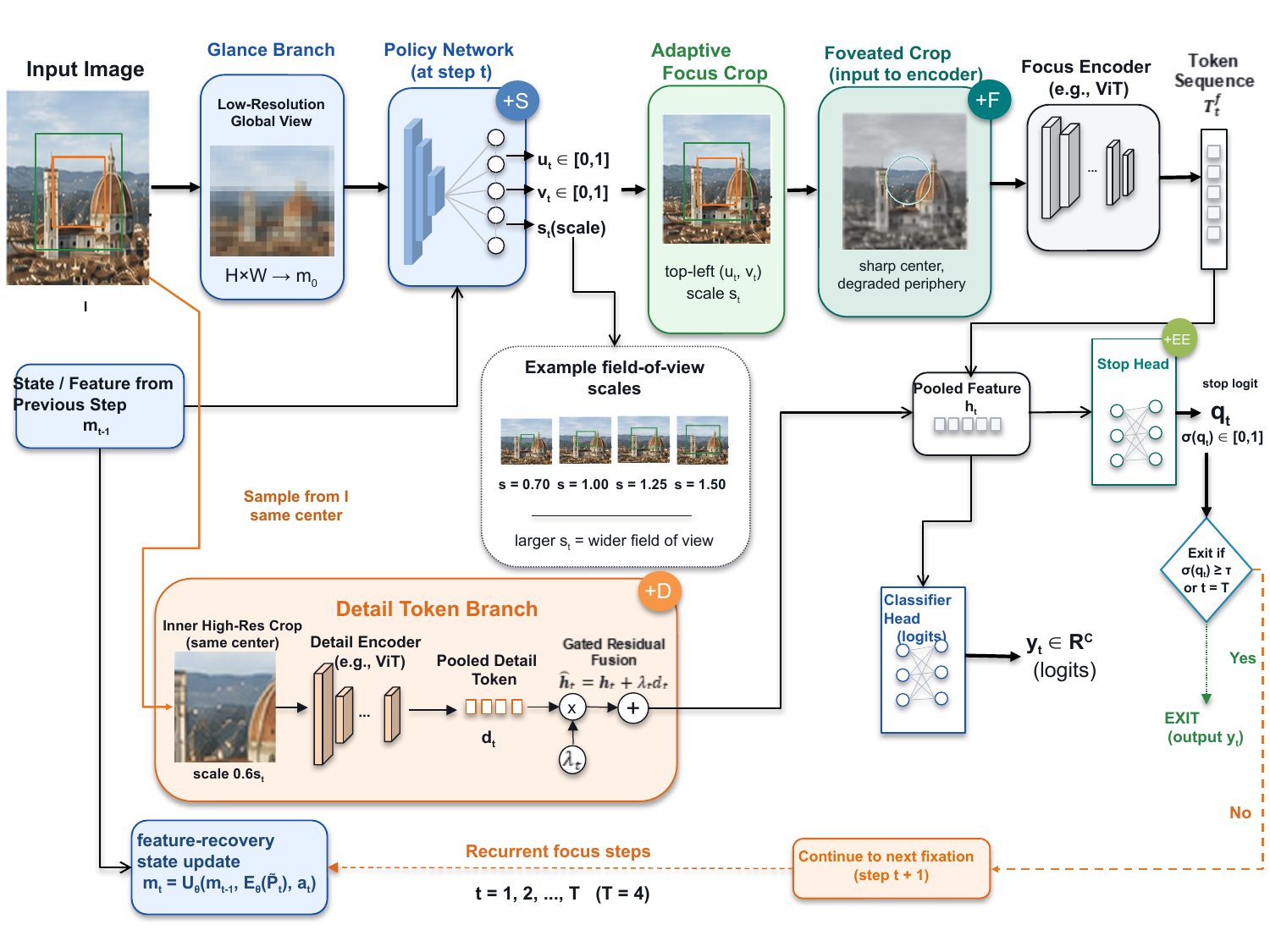}
  \caption{Overview of the controllable observation loop. A low-resolution glance and the previous-step state drive a policy that predicts crop location and field-of-view scale (+S). The selected crop is optionally foveated (+F), encoded by the shared focus branch, augmented by an inner high-resolution detail-token branch (+D), and routed to classifier and label-grounded stop heads (+EE). Larger $s_t$ gives a wider field of view. Here, $(u_t,v_t)$ denotes the normalized top-left crop position, $q_t$ is the stop logit, $\sigma$ is the sigmoid function, and $\tau$ is the exit threshold. The dashed path continues only when the stopping score $\sigma(q_t)<\tau$ and $t<T$, where $T$ is the final focus step. The inner detail crop is sampled directly from $I$ at the outer crop's center, with scale $0.6s_t$. Green and orange boxes illustrate the outer and inner crop regions, respectively.}
  \label{fig:pipeline}
\end{figure}

We introduce a scale-adaptive extension of AdaptiveNN, summarized in \cref{fig:pipeline}. The policy predicts location and scale for each fixation, allowing the model to zoom in on discriminative regions or zoom out for broader context. Selected centers are revisited by a residual detail-token path at higher resolution, with optional scale-gated foveation that suppresses peripheral detail in expanded views. A label-grounded stop head maps the current representation to a scalar stop logit $q_t$. At inference time, the model stops at the first focus step whose score $\sigma(q_t)$ reaches a chosen threshold $\tau$, or at the final step if the threshold is never reached. Varying $\tau$ changes the stopping criterion without retraining the model, allowing us to evaluate the resulting accuracy-compute trade-off on each dataset.

Our contributions are:
\begin{itemize}
  \item a controllable observation interface for active recognition that unifies geometry, evidence, and stopping control on one sequential state in the public AdaptiveNN-DeiT-S pipeline. It is implemented as a scale-adaptive fixation action, residual detail-token fusion with optional scale-gated foveation, and label-grounded per-step stop heads;
  \item an evaluation protocol that separates realized sequential exits from simulated stop-rule diagnostics, anchors component effects to a matched-budget same-codebase baseline, and compares against our reruns of the unmodified AdaptiveNN implementation released by its authors; and
  \item an empirical analysis across fine-grained datasets, full ImageNet-1K, and ImageNet-A/R-derived classification splits that identifies where scale and detail improve accuracy and where thresholded stopping trades computation for accuracy.
\end{itemize}

\section{Related Work}
\label{sec:related}

\textbf{Adaptive recognition and attention.}
Vision Transformers~\cite{dosovitskiy2021vit} and DeiT~\cite{touvron2021deit} are strong fixed-computation backbones, whereas AdaptiveNN~\cite{wang2025adaptivenn} casts recognition as sequential decision-making that chooses focus regions after an initial glance, in the spirit of recurrent attention models~\cite{mnih2014ram} and spatial transformers~\cite{jaderberg2015stn} that learn where to sample. We retain this coarse-to-fine structure and extend its action space and inference policy.

\textbf{Foveated and multi-scale computation.}
Foveated vision allocates high fidelity to the gaze center and less to the periphery, as studied in foveal-peripheral networks~\cite{deza2020emergent,lukanov2021fovealperipheral}, foveated rendering and magnification~\cite{patney2016foveated,thavamani2021fovea}, and active recognition systems~\cite{jonnalagadda2021foveater,killick2023foveation}. Fine-grained categorization~\cite{maji2013aircraft,krause2013cars,wah2011cub}, which our fixation policy targets, commonly relies on part features~\cite{zhang2014partbased}, attention~\cite{fu2017lookclose}, or feature pyramids~\cite{lin2017fpn} rather than gaze-centered sensing. We treat foveation as a scale-gated crop refinement rather than a standalone sensor or fixed pyramid, and our continuous scale coordinate selects one adaptive field of view per fixation, optimized by the policy reward.

Recent work also studies non-uniform visual representations under limited pixel budgets. Seeing More with Less~\cite{gizdov2025seeing} evaluates single-fixation foveated sampling for visual question answering and object detection. LLMind~\cite{debnath2026llmind} adapts M\"obius-parameterized sampling through test-time semantic feedback from a frozen vision-language model. Our method addresses sequential image recognition: a trained fixation policy uses the evolving state to choose the next crop's location and scale, while a stop head determines whether to acquire another observation.

\textbf{Early exit and adaptive inference.}
Early-exit networks attach auxiliary classifiers or stopping policies to intermediate representations so that inference terminates early for easy examples~\cite{teerapittayanon2016branchynet,huang2018msdnet,elbayad2020depthadaptive}, and dynamic vision models prune or halt tokens within a fixed view, adapt depth, width, and routing, or cascade input resolutions per image~\cite{rao2021dynamicvit,yin2022avit,wang2021dvt,han2022dynamicsurvey}. These decisions act along network depth or token count within a fixed passive view, under heterogeneous backbones, resolutions, and recipes, so their accuracy-compute frontiers are not directly comparable to an active observation loop. We therefore evaluate controllability within one public active-recognition pipeline. We build on the glance-and-focus paradigm for active recognition~\cite{wang2020glancefocus,wang2025adaptivenn}, which stops once the classifier passes a confidence threshold; our label-grounded stop heads instead interact with the learned per-fixation scale and provide multiple thresholds from the same checkpoint.

\section{Method}
\label{sec:method}

\textbf{Overall framework.}
The method builds on the AdaptiveNN-DeiT-S~\cite{wang2025adaptivenn} glance-and-focus structure and extends its fixed four-step focus sequence into an adaptive observation controller. A low-resolution glance initializes a spatial token state. At each focus step, the policy predicts a crop location and field-of-view scale; the selected crop is transformed by scale-gated foveation, encoded by the focus branch, optionally complemented by an inner high-resolution detail descriptor, and then classified. A per-step stop head maps the same representation to a stopping score. The trained model therefore controls three quantities at inference time: field of view, local evidence quality, and observation count.

\subsection{Controllable Observation Loop}

Throughout the paper, we index focus steps starting from 1, with the glance prediction denoted as step 0. Let $m_0$ denote the spatial token state produced by the glance branch. At focus step $t$, the policy reads the detached state $m_{t-1}$ and predicts an action
\begin{equation}
  a_t \sim \pi_\psi(\cdot \mid m_{t-1}), \qquad
  a_t=(u_t,v_t,z_t).
  \label{eq:policy_action}
\end{equation}
The first two coordinates select the normalized top-left crop position, and the third coordinate controls the field of view. The observation path can be summarized as
\begin{equation}
  \begin{aligned}
  P_t &= \mathrm{Crop}(I;u_t,v_t,s_t),\\
  \widetilde{P}_t &= \mathrm{Fovea}(P_t;s_t,t,e),\\
  m_t &= U_\theta(m_{t-1},E_\theta(\widetilde{P}_t),a_t),
  \end{aligned}
  \label{eq:loop_update}
\end{equation}
where $E_\theta$ is the focus encoder, $U_\theta$ is the feature-recovery update, $e$ is the training epoch, and $\widetilde{P}_t=P_t$ when foveation is disabled or has zero strength. A pooled classifier feature $h_t$ is obtained from $m_t$. When enabled, the detail path adds a residual descriptor to $h_t$ before the classifier and stop heads produce $y_t$ and $q_t$. The exit step under threshold $\tau$ is
\begin{equation}
  t^\star = \min\{t:\sigma(q_t)\ge\tau\},
  \label{eq:exit_step}
\end{equation}
with $t^\star=T$ if no earlier step satisfies the condition.

\subsection{Baseline AdaptiveNN}

Given image $I$, a glance branch processes a low-resolution global view and produces the spatial token state used by the fixation policy. At focus step $t \in \{1,\ldots,T\}$, the policy predicts a fixation location, the focus branch encodes the selected patch, the feature-recovery module writes the local evidence back into the spatial state, and a classifier predicts logits $y_t$. We use the default four-step setting ($T=4$). The original policy outputs two continuous action components,
\begin{equation}
  a_t = (u_t, v_t), \qquad u_t,v_t \in [0,1],
\end{equation}
which define the top-left crop coordinate after scaling to the image extent.

Training combines supervised classification losses on the glance and focus predictions, knowledge distillation from the final focus prediction to earlier predictions, and a PPO~\cite{schulman2017ppo} update for the non-differentiable fixation policy. We retain this optimization structure, modifying only the action space, observation transforms, and stopping head described below.

\subsection{Scale-Adaptive Fixations}

The baseline policy chooses only \emph{where} to look: the crop size is fixed, so the field of view cannot adapt to objects of different apparent size. We extend the action space from two-dimensional location to location plus scale,
\begin{equation}
  a_t = (u_t, v_t, z_t), \qquad
  s_t = s_{\min} + (s_{\max} - s_{\min}) z_t,
  \label{eq:scale}
\end{equation}
where $z_t=\sigma(\ell_t)\in[0,1]$ is produced by a third policy output and $s_t$ is the relative field-of-view scale. The extracted crop has side length proportional to $s_t$ and is resized to the fixed focus input size, so the encoder cost per focus step is unchanged by the chosen scale. When the scale head is disabled or fixed at $s_t=1$, the model recovers the original fixed-field-of-view behavior. We initialize the scale-head weight to zero and its bias to
\begin{equation}
  \mathrm{logit}\left(\frac{s_{\mathrm{anchor}}-s_{\min}}{s_{\max}-s_{\min}}\right),
  \qquad s_{\mathrm{anchor}}=1.0,
  \label{eq:scale_init}
\end{equation}
so training starts from the baseline field of view. The scale coordinate shares the same PPO objective as the location coordinates. Because the reward is unchanged, scale is learned from recognition improvements rather than from an auxiliary scale loss.

For the fine-grained datasets and ImageNet-A we use $s_{\min}=0.70$ and $s_{\max}=1.50$, allowing the policy to zoom in on small discriminative parts or zoom out for broader context; for ImageNet-1K and ImageNet-R we use the narrower range $[0.75,1.35]$ to limit scale variation on datasets with more uniform object sizes.

\subsection{Foveated Focus Crop}

We use foveation as a scale-gated transform of the selected focus crop rather than as a standalone sensor. Given the focus crop $P_t$ at scale $s_t$, the model preserves the central region and blends the crop with a degraded version of itself:
\begin{equation}
  \widetilde{P}_t = (1-\alpha_t)P_t + \alpha_t\left(M \odot P_t + (1-M) \odot D(P_t)\right),
  \label{eq:fovea}
\end{equation}
where $M$ is a radial center mask and $D(\cdot)$ downsamples the crop to 50\% of its side length before upsampling it back. The mask is 1 inside radius 0.35, decreases linearly across a 0.10 blending band centered at radius 0.40, and is 0 outside radius 0.45 in normalized crop coordinates. This implements the central-versus-peripheral fidelity trade-off used in foveated vision models~\cite{deza2020emergent,lukanov2021fovealperipheral,killick2023foveation}; when $\alpha_t=0$, the operation reduces to the original crop.

The effective foveation strength is scale-gated:
\begin{equation}
  \alpha_t = \alpha_{\max} r(e)\,\mathbf{1}[t \ge t_0]\,
  \mathrm{clip}\left(\frac{s_t - s_{\mathrm{anchor}}}{s_{\max} - s_{\mathrm{anchor}}}, 0, 1\right),
  \label{eq:fovea_gate}
\end{equation}
where $\alpha_{\max}=0.5$, $s_{\mathrm{anchor}}=1.0$, $s_{\max}$ is the dataset-specific maximum scale from \cref{eq:scale}, $t_0=2$ under the step indexing defined above, and $r(e)$ is the training ramp
\begin{equation}
  r(e)=0\ (e<10), \qquad
  r(e)=\frac{e-10}{10}\ (10\le e<20), \qquad
  r(e)=1\ (e\ge20).
  \label{eq:fovea_ramp}
\end{equation}
Early-exit fine-tuning starts from the trained +S+F+D checkpoint with $r(e)=1$. The scale gate makes foveation strongest for expanded views and nearly inactive near the anchor scale, acting as a conditional refinement of the evidence path whose dataset-dependent effect is quantified in \cref{sec:results}.

\subsection{Detail-Token Fusion}

Foveation controls peripheral fidelity, but fine-grained evidence can still be under-represented in the selected center. We therefore use a detail branch that extracts an inner crop aligned to the center of the selected outer crop. The inner crop is sampled from the original image using the same fixation center as $P_t$, with scale $0.6s_t$ in the main experiments, giving a higher-resolution view of the innermost region.

Because the inner crop shares the outer crop's center, the policy makes one location decision for both paths. The detail branch then samples a smaller region from the original image, rather than magnifying the resized outer crop. This distinction preserves access to image detail that outer-crop resizing may discard. The fixed scale ratio also lets scale control change the two fields of view together.

The inner crop is processed by the shared focus encoder, and its features are pooled around the spatial center into a descriptor $p_t$. Scale-gated foveation therefore modulates peripheral fidelity within the main focus crop, whereas the detail path injects a separate high-resolution central descriptor. In the reported setting, the descriptor is the average of the central $3\times3$ token window after the focus encoder. We detach the inner features before the detail gate so that gradients from the auxiliary detail path do not overwrite the main focus encoder updates:
\begin{equation}
  p_t = \mathrm{CenterPool}\!\left(\mathrm{sg}\!\left[E_\theta(P_t^{\mathrm{inner}})\right]\right), \qquad
  d_t = \tanh(G_\phi(p_t)), \qquad
  \widehat{h}_t = h_t + \lambda_t d_t,
  \label{eq:detail}
\end{equation}
where $\mathrm{sg}[\cdot]$ denotes stop-gradient, $G_\phi$ is a two-layer MLP with hidden width $2C$ and GELU activation, $C$ is the focus feature dimension, and $\lambda_t$ ramps from 0 to 0.5 over epochs 10--18 in non-exit runs. The main experiments use one pooled detail token, center pooling, residual fusion, and no write-back into the next policy state. The detail branch adds one focus-encoder pass per detail-enabled executed step; we enable detail from the second focus step onward. Unlike recurrent attention CNNs that stack additional sub-networks for magnified regions~\cite{fu2017lookclose}, our design reuses the shared focus encoder and injects only one stop-gradient pooled token. For conservative accounting, the supplementary encoder-FLOP estimate counts a detail pass at every executed focus step, including the first.

\subsection{Confidence-Conditioned Early Exit}

The full model adds a scalar stop head to each focus step in the four-step observation path. The stop head consumes the pooled feature before classification and outputs a stop logit $q_t$. At evaluation, the model exits at the first step satisfying
\begin{equation}
  \sigma(q_t) \ge \tau,
\end{equation}
or at the final step $T=4$ if no earlier step passes the threshold. The threshold $\tau$ is an inference-time hyperparameter, distinct from the confidence threshold used to construct training labels. Since the reported early-exit runs use hard class labels (no mixup), we use a hard-label stop target. Let $\rho_t=\mathrm{softmax}(y_t)$, $\widehat{c}_t=\max_c \rho_t(c)$, $\widehat{y}_t=\arg\max_c \rho_t(c)$, and $y^\star$ be the ground-truth class. The stop target is
\begin{equation}
  b_t =
  \mathbf{1}\!\left[\widehat{y}_t = y^\star\right]\,
  \mathbf{1}\!\left[\widehat{c}_t \ge 0.7\right].
  \label{eq:stop_target}
\end{equation}
This label-grounded target trains the stop head against task correctness rather than against agreement with the final step.

At inference, the stopping rule uses only the learned score and does not require ground-truth labels. The confidence threshold in the training target therefore does not replace the inference threshold $\tau$. Changing $\tau$ selects an intermediate prediction without altering its logits. Because later observations may change the predicted class, neither the stopping score nor the final-step accuracy establishes the accuracy of the returned predictions; this must be evaluated at the selected exits.

The stop head is trained with binary cross-entropy,
\begin{equation}
  \mathcal{L}_{\mathrm{stop}} =
  \frac{1}{T}\sum_{t=1}^{T}
  \mathrm{BCEWithLogits}(q_t,b_t),
  \label{eq:stop_loss}
\end{equation}
added to the main classification loss with auxiliary weight 0.05. Early-exit fine-tuning initializes from the +S+F+D checkpoint and runs for 15--20 additional epochs (dataset schedules in the supplementary material). At inference, varying $\tau$ traces out a range of accuracy-compute operating points from the same checkpoint.

Because a stop decision changes how many observations are executed per sample, we report two protocols separately. In \emph{simulation} mode the model runs the full four-step pass and applies the stop rule post hoc, analyzing the policy without claiming realized savings. In \emph{sequential-step} mode samples terminate when the threshold fires, and the realized step count is measured. Since $y_t$ and $q_t$ are produced before any later step, gathering canonical step-$t$ logits from a full unrolled pass introduces no temporal leakage; we use these for protocol-comparable accuracy at the measured exit steps and additionally record physically truncated logits for consistency checks. Only sequential-step rows support realized compute-saving claims.

We report accuracy and focus-step ratio (average selected focus steps divided by $T=4$); the supplementary cost breakdown translates this ratio into a conservative encoder-FLOP estimate, including the always-on glance overhead and excluding crop sampling and other non-encoder operations.

\subsection{Training Objective}

Training separates differentiable prediction losses from the non-differentiable fixation policy update. The backbone, classifiers, detail branch, and stop heads are optimized with
\begin{equation}
  \mathcal{L}_{\mathrm{sup}}
  = \mathcal{L}_{\mathrm{cls}}
  + \lambda_{\mathrm{kd}}\mathcal{L}_{\mathrm{kd}}
  + \lambda_{\mathrm{stop}}\mathcal{L}_{\mathrm{stop}},
  \label{eq:training_objective}
\end{equation}
where $\mathcal{L}_{\mathrm{cls}}$ is the sum of supervised cross-entropy losses on the glance prediction, focus predictions, and focus-only auxiliary prediction, and $\mathcal{L}_{\mathrm{kd}}$ distills the final focus prediction to the glance and earlier focus predictions using temperature 1. We set $\lambda_{\mathrm{kd}}=1$. The stop term is enabled only for early-exit fine-tuning; for these runs, $\lambda_{\mathrm{stop}}=0.05$, and otherwise $\lambda_{\mathrm{stop}}=0$.

The fixation policy is updated separately with PPO because crop selection is non-differentiable. For each sample and focus step, the reward is defined as the reduction in cross entropy from the previous prediction to the current focus prediction:
\begin{equation}
  R_t = \mathrm{CE}_{t-1} - \mathrm{CE}_{t},
  \label{eq:ppo_reward}
\end{equation}
where $\mathrm{CE}_{0}$ is computed from the glance prediction and $\mathrm{CE}_{t}$ from focus step $t$ after the enabled observation transforms and detail fusion. The learned scale coordinate shares this policy reward, with no auxiliary scale regression loss or scale-cost penalty.

The components are trained in stages under a shared recipe. All non-exit variants start from the same AdaptiveNN checkpoint. During scale warmup, the scale coordinate in both the executed action and PPO latent action is fixed to the anchor so the crop field of view stays at the baseline; after warmup, scale is sampled and optimized with location. Foveation and detail fusion use the ramps in \cref{eq:fovea_ramp,eq:detail}. Early-exit fine-tuning starts from the trained +S+F+D checkpoint and adds only the stop supervision in \cref{eq:stop_target,eq:stop_loss}. Dataset-specific schedules are kept in the supplementary material.

\section{Experiments}
\label{sec:experiments}

\subsection{Datasets and Protocol}

We evaluate on eight recognition benchmarks: FGVC-Aircraft~\cite{maji2013aircraft}, Stanford Cars~\cite{krause2013cars}, CUB-200-2011~\cite{wah2011cub}, Stanford Dogs~\cite{khosla2011dogs}, Oxford-IIIT Pet~\cite{parkhi2012pets}, ImageNet-A~\cite{hendrycks2021natural}, ImageNet-R~\cite{hendrycks2021many}, and the full ImageNet-1K validation set~\cite{deng2009imagenet}, covering part-sensitive fine-grained recognition, ImageNet-A/R-derived classification splits, and large-vocabulary recognition. Following the AdaptiveNN protocol, we report accuracy on each dataset's public evaluation split, which we call validation throughout even when a release names it ``test''.

All non-exit variants initialize from the same AdaptiveNN checkpoint and fine-tune DeiT-S glance and focus branches for four focus steps ($T=4$) with AdamW~\cite{loshchilov2019adamw}, cosine decay, RandAugment~\cite{cubuk2020randaugment}, and PPO updates with incremental cross-entropy reward; Full+EE initializes from the trained +S+F+D checkpoint. Dataset-specific schedules are summarized in the supplementary material. We select checkpoints by best public-validation Acc@1 and report Acc@5 for context. Evaluation uses \texttt{crop\_pct}=1.0 with input size 336 for Aircraft, Cars, CUB, Dogs, Oxford-Pet, and ImageNet-A, and 288 for ImageNet-1K and ImageNet-R; ImageNet-R uses a fixed 10\% validation split (seed 0), matching the split used during training.

\subsection{Compared Variants}

We compare our AdaptiveNN rerun against additive variants:
\begin{itemize}
  \item \textbf{Orig}: the original AdaptiveNN configuration, rerun \emph{inside our modified codebase} with all new components disabled. It uses fixed-scale focus crops and no foveation, detail fusion, or stop head, but shares the training recipe used by the variants below;
  \item \textbf{+S}: learned scale action;
  \item \textbf{+S+F}: learned scale with foveated focus crops;
  \item \textbf{+S+D}: learned scale with detail-token fusion;
  \item \textbf{+S+F+D}: the non-exit full adaptive crop model; and
  \item \textbf{Full+EE}: +S+F+D with the confidence-conditioned stop head.
\end{itemize}
For Aircraft, Cars, CUB, Dogs, ImageNet-A, ImageNet-1K, ImageNet-R, and Oxford-Pet we also rerun the \emph{unmodified official AdaptiveNN codebase}~\cite{wang2025adaptivenn} (``Official'' in \cref{tab:main}) as the external fixed-step reference. \textbf{Orig} remains the internal same-codebase anchor for additive component effects in \cref{tab:ablation}; the supplementary material reports its difference from Official separately.

\section{Results}
\label{sec:results}

We first compare accuracy against the strongest available fixed-step baselines (\cref{tab:main}; budget-matched diagnostic in the supplementary material), then isolate the geometry and evidence controls (\cref{tab:ablation}, with supplementary scale statistics), and finally evaluate stopping as a thresholded accuracy-compute trade-off (\cref{tab:earlyexit}, \cref{fig:frontier_seq}).

\subsection{Accuracy and Budget Diagnostics}

\Cref{tab:main} summarizes the main validation results for the scale-adaptive detail model. To keep the main comparison compact, we show the external \textbf{Official} rerun and the strongest completed in-codebase non-exit row, selected from Orig and the four additive non-exit variants. Our primary metric, $\Delta$best, subtracts the stronger of these fixed/non-exit references from Full+EE. Fine-grained rows report the seed-0 run and the ImageNet-A/-1K/-R rows are single runs; three-seed paired checks are summarized below. Accuracy rows are not compute-neutral---a full four-step Full+EE unroll costs 9.90 versus 5.50 encoder-pass GFLOPs for Orig (supplementary cost breakdown)---so realized savings arise only under sequential early exit (\cref{tab:earlyexit}). The supplementary material reports the Orig--Official gap and the budget-matched continuation diagnostic.

Full+EE exceeds the official rerun on seven of eight datasets and is within 0.002 points on ImageNet-1K. The only negative $\Delta$best values occur on ImageNet-1K ($-0.158$) and ImageNet-R ($-0.267$), where +S+D is stronger. Small margins below 0.1 points are treated as parity-level differences.

The additional training stage also affects how improvements should be attributed. Full+EE receives further fine-tuning, so its difference from an original-schedule non-exit model combines the stopping objective with additional optimization. The continuation diagnostic in the supplementary material compares models after matched additional training to assess this source of variation. Paired ImageNet-1K tests quantify uncertainty over aligned validation examples for fixed checkpoints, whereas the repeated-seed studies examine variation across independently trained models. These analyses answer different questions. A single-run ranking alone therefore does not establish that one configuration reliably outperforms another across training seeds.

\begin{table}[tb]
  \caption{Main validation accuracy (Acc@1/Acc@5 at the selected 1-based epoch). $\Delta$best is Full+EE minus the stronger of the Official rerun and the best in-codebase non-exit variant (selected row named in parentheses). $^\dagger$+S+F also achieves 66.138\% Acc@1 on ImageNet-A. The highest Acc@1 in each row is shown in bold; Table~\ref{tab:ablation} uses the same convention.}
  \label{tab:main}
  \centering
  \scriptsize
  \setlength{\tabcolsep}{4.2pt}
  \resizebox{\linewidth}{!}{%
  \begin{tabular}{lcccc}
    \toprule
    Dataset & Official & Best in-codebase non-exit & Full+EE (ours) & $\Delta$best \\
    \midrule
    Aircraft & 77.318/97.000 @75 & 77.078/96.550 @69 (+S+F+D) & \best{77.558}/96.910 @10 & \gain{0.240} \\
    Cars & 83.076/97.525 @79 & 82.741/97.675 @77 (+S+F+D) & \best{83.711}/97.650 @15 & \gain{0.635} \\
    CUB & 84.524/96.912 @65 & 84.369/96.929 @79 (Orig) & \best{84.800}/97.084 @17 & \gain{0.276} \\
    Dogs & 91.981/99.452 @30 & 92.401/99.464 @22 (+S+D) & \best{92.413}/99.534 @3 & \gain{0.012} \\
    ImageNet-1K & 82.126/95.946 @1 & \best{82.282}/96.014 @1 (+S+D) & 82.124/95.896 @5 & $-0.158$ \\
    ImageNet-A & 64.815/80.688 @46 & 66.138/82.011 @42 (+S+D)$^\dagger$ & \best{66.534}/81.746 @17 & \gain{0.396} \\
    ImageNet-R & 85.133/94.500 @37 & \best{85.900}/95.033 @32 (+S+D) & 85.633/94.933 @14 & $-0.267$ \\
    Oxford-Pet & 96.402/99.838 @68 & 96.537/99.865 @66 (Orig) & \best{96.591}/99.919 @20 & \gain{0.054} \\
    \bottomrule
  \end{tabular}}
\end{table}

\subsection{Geometry and Evidence Controls}

\Cref{tab:ablation} gives the component ablation behind the dataset-dependent accuracy behavior. We read +S, +D, and +F as coupled controls: scale action sets field-of-view geometry, detail-token fusion injects higher-resolution local evidence, and foveation re-weights periphery with strength gated by the learned scale through \cref{eq:fovea_gate}. 

Scale alone improves Acc@1 on Cars (+0.560), Dogs (+0.327), Aircraft (+0.390), and ImageNet-1K (+0.118). Among the non-exit variants with added components, +S+D or +S+F+D achieves the highest Acc@1 on each benchmark: +S+D is strongest on Dogs, ImageNet-1K, ImageNet-R, and Oxford-Pet and ties +S+F on ImageNet-A, while +S+F+D is strongest on Aircraft, Cars, and CUB. 

Foveation is activated through the scale gate and is not uniformly helpful. For the Full+EE checkpoints, the supplementary statistics show mean scales near the anchor on Aircraft, CUB, Dogs, and ImageNet-A; these means do not determine the strength of foveation at individual fixations. In the non-exit ablation, +S+F performs close to +S on Aircraft, CUB, and Dogs, whereas ImageNet-A has a 0.397-point difference in the single-run comparison. Cars benefits from +S+F+D, while ImageNet-1K, ImageNet-R, and Oxford-Pet favor +S+D over +S+F+D. Thus, the component comparisons show a dataset-dependent effect rather than a uniform benefit from foveation.

\begin{figure}[tb]
  \centering
  \includegraphics[width=1\linewidth]{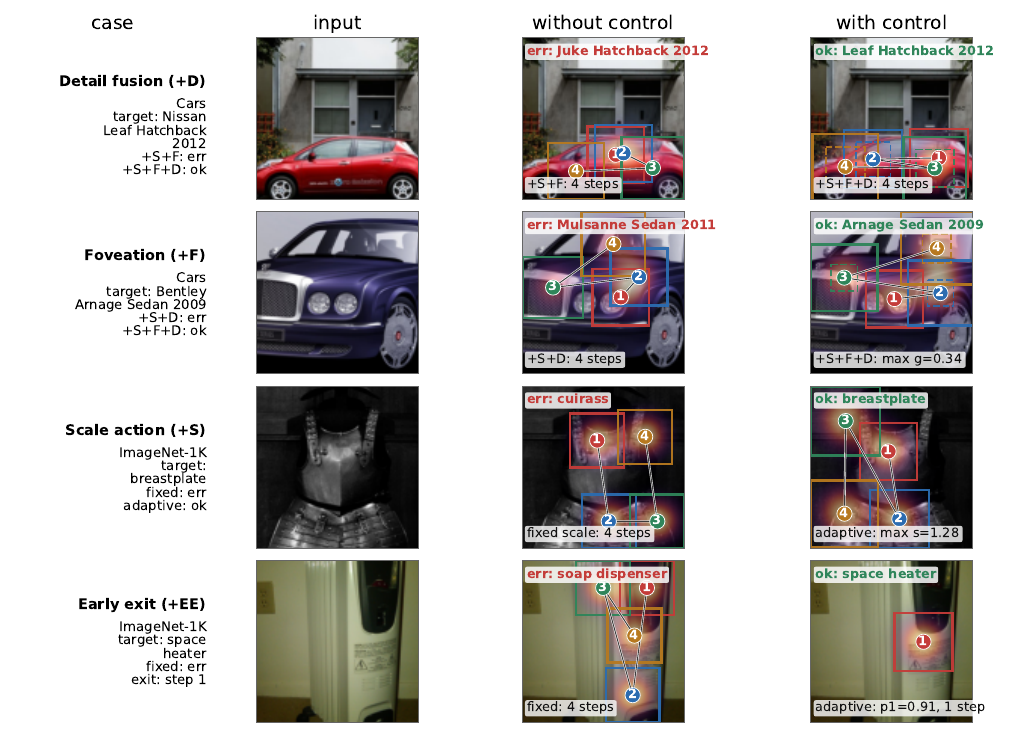}
  \caption{Qualitative module contrasts on held-out examples. The Cars rows compare detail fusion (+S+F to +S+F+D) and scale-gated foveation (+S+D to +S+F+D). The ImageNet-1K rows compare Orig with Full+EE as joint-model contrasts rather than controlled, single-component comparisons. Colored boxes and numbers mark focus order; the heat overlay is per-example fixation density; dashed inner boxes mark detail crops (top row) and the sharp foveal core (second row) from step 2 onward.}
  \label{fig:qualitative_fixations}
\end{figure}

\Cref{fig:qualitative_fixations} illustrates the detail, foveation, scale, and stopping controls at the example level.

\begin{table}[tb]
  \caption{Acc@1 ablation across additive model variants. +S, +F, and +D denote learned scale, scale-gated foveation, and detail-token fusion, respectively; Full+EE adds early-exit fine-tuning on top of +S+F+D.}
  \label{tab:ablation}
  \centering
  \scriptsize
  \setlength{\tabcolsep}{3.4pt}
  \resizebox{\linewidth}{!}{%
  \begin{tabular}{lrrrrrr}
    \toprule
    Dataset & Orig & +S & +S+F & +S+D & +S+F+D & Full+EE \\
    \midrule
    Aircraft & 76.358 & 76.748 & 76.898 & 76.628 & 77.078 & \best{77.558} \\
    Cars & 81.696 & 82.256 & 81.771 & 82.467 & 82.741 & \best{83.711} \\
    CUB & 84.369 & 84.179 & 84.300 & 84.300 & 84.334 & \best{84.800} \\
    Dogs & 92.051 & 92.378 & 92.238 & 92.401 & 92.238 & \best{92.413} \\
    ImageNet-1K & 82.116 & 82.234 & 82.146 & \best{82.282} & 82.246 & 82.124 \\
    ImageNet-A & 65.873 & 65.741 & 66.138 & 66.138 & 65.873 & \best{66.534} \\
    ImageNet-R & 85.467 & 85.300 & 85.400 & \best{85.900} & 85.133 & 85.633 \\
    Oxford-Pet & 96.537 & 96.402 & 96.429 & 96.456 & 96.429 & \best{96.591} \\
    \bottomrule
  \end{tabular}}
\end{table}

\FloatBarrier
\subsection{Stopping-Control Trade-Off}

\Cref{tab:earlyexit} reports confidence-conditioned stopping at thresholds $\tau\in\{0.70,0.90\}$. Sequential-step rows measure realized exits; simulation rows apply the stopping rule to full rollouts and do not establish realized savings. The supplementary material reports the complete threshold sweep and encoder-pass estimates.

\begin{table}[tb]
  \caption{Confidence-conditioned early exit at $\tau\in\{0.70,0.90\}$. \textbf{Seq. steps}: realized exits (Acc@1 from canonical logits at the exit step; steps measured and encoder FLOPs estimated from those steps). \textbf{Sim.}: diagnostic stop rule on full four-step outputs only; these rows do not measure realized inference cost or FLOP savings. Ratio is average focus steps over four. Full is the full-unroll Acc@1 of the same checkpoint measured in the corresponding protocol environment; for ImageNet-1K this is the sequential evaluation, hence 82.040 versus the main-table 82.124.}
  \label{tab:earlyexit}
  \centering
  \scriptsize
  \setlength{\tabcolsep}{2.8pt}
  \resizebox{\linewidth}{!}{%
  \begin{tabular}{llrrrrrrr}
    \toprule
    & & & \multicolumn{3}{c}{$\tau=0.70$} & \multicolumn{3}{c}{$\tau=0.90$} \\
    \cmidrule(lr){4-6} \cmidrule(lr){7-9}
    Dataset & Protocol & Full & Acc@1 & Avg. step & Ratio & Acc@1 & Avg. step & Ratio \\
    \midrule
    Aircraft & Sim. & 77.558 & 77.378 & 3.669 & 0.917 & 77.558 & 4.000 & 1.000 \\
    Cars & Sim. & 83.711 & 83.686 & 3.988 & 0.997 & 83.711 & 4.000 & 1.000 \\
    CUB & Sim. & 84.800 & 84.800 & 3.710 & 0.927 & 84.800 & 4.000 & 1.000 \\
    Dogs & Sim. & 92.413 & 92.424 & 3.992 & 0.998 & 92.413 & 4.000 & 1.000 \\
    ImageNet-1K & Seq. steps & 82.040 & 81.404 & 1.731 & 0.433 & 81.888 & 2.446 & 0.611 \\
    ImageNet-A & Sim. & 66.534 & 66.534 & 3.959 & 0.990 & 66.534 & 4.000 & 1.000 \\
    ImageNet-R & Sim. & 85.633 & 84.333 & 1.437 & 0.359 & 85.033 & 1.926 & 0.481 \\
    Oxford-Pet & Seq. steps & 96.591 & 96.050 & 1.343 & 0.336 & 96.591 & 4.000 & 1.000 \\
    \bottomrule
  \end{tabular}}
\end{table}

\begin{figure}[tb]
  \centering
  \includegraphics[width=1\linewidth]{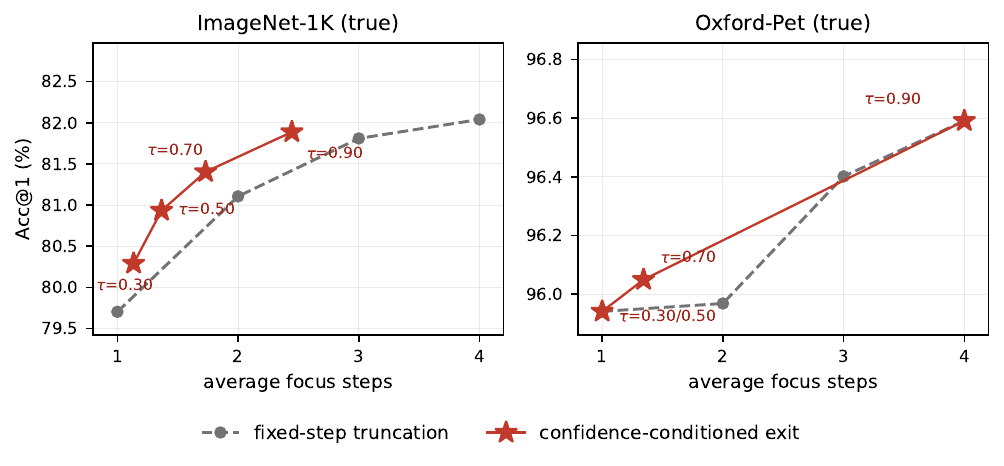}
  \caption{Realized sequential accuracy versus average focus steps for the Full+EE checkpoint. Red markers are realized confidence-conditioned exits from one checkpoint with measured average focus steps; dashed gray curves are fixed-step truncation at integer budgets. The supplementary material reports the full mixed diagnostic, including ImageNet-R simulation-only points.}
  \label{fig:frontier_seq}
\end{figure}

\Cref{fig:frontier_seq} compares confidence-conditioned exits with fixed-step truncation of the same checkpoint under realized sequential execution. On ImageNet-1K the sequential $\tau=0.70$ point outperforms fixed two-step truncation by 0.298 percentage points in Acc@1 while using fewer than two focus steps on average, and $\tau=0.90$ outperforms fixed three-step truncation in Acc@1 while using fewer focus steps on average. Oxford-Pet shows a low-cost $\tau=0.70$ operating point and conservatively returns to full-step behavior at $\tau=0.90$. The stop head thus reallocates observations across examples, with realized savings supported only by the sequential-step rows of \cref{tab:earlyexit}.

The cost comparison depends on the reference model and the selected step count. For example, the ImageNet-1K $\tau=0.90$ setting uses about 2.45 focus steps but costs 6.48 encoder-pass GFLOPs under the conservative supplementary estimate, compared with 5.50 for four-step Orig. It thus saves computation relative to four-step Full+EE while remaining above Orig under this proxy. The supplementary runtime study uses a separately fitted max-softmax stopping rule and measures model-forward execution. Its timings characterize that profiling setup and should not be assigned directly to the learned-head threshold points in \cref{tab:earlyexit}. Step reduction, encoder cost, and measured latency describe related but distinct quantities.

\section{Limitations}
\label{sec:limitations}

\textbf{Evaluation scope.} Three-seed studies cover Cars, Aircraft, CUB, Dogs, and Oxford-Pet; the original ImageNet-A, ImageNet-1K, and ImageNet-R table rows use single runs. A separate single-GPU study in the supplementary material examines three training seeds for ImageNet-R. Checkpoints are selected by best public-split Acc@1, so parity-level margins are sensitive to epoch selection (best-versus-last table in the supplementary material).

\textbf{Accuracy and cost.} Extra training of the fixed-step model is itself a strong baseline, so matched-budget gains are dataset-dependent. Detail-token fusion adds one focus-encoder pass from the second focus step onward, so rarely-exiting datasets incur most of the additional cost; the supplementary runtime measurements cover ImageNet-1K model-forward execution on one GPU.

\textbf{Component and threshold generality.} Interactions are not uniform---ImageNet-R favors +S+D over active foveation---and the thresholds in \cref{tab:earlyexit} are set manually.

\section{Conclusion}
\label{sec:conclusion}

\looseness=-1 We extended a fixed-step AdaptiveNN recognizer with a \emph{controllable observation interface}: learned scale controls field of view, detail-token fusion with optional scale-gated foveation controls local evidence, and confidence-conditioned stop heads control observation count. Our method shows competitive performance across benchmark datasets while providing a threshold-controlled accuracy-compute frontier from one checkpoint: the measured sequential ImageNet-1K run sustains 81.9\% top-1 at $\tau=0.90$ using 2.45 focus steps. Since the detail path adds per-step cost, net compute benefit depends on realized early termination. The resulting framework offers explicit, tunable control over observation geometry, evidence, and stopping.

% ---- Bibliography ----
\bibliographystyle{splncs04}
\bibliography{main}

\end{document}

\title{Adaptive Observation Control for Active Visual Recognition\\Supplementary Material}
\titlerunning{Supplementary Material}

% Shared publication metadata for the main paper and supplement.
% The shared star identifies Weijia Wang and Yuan-Gen Wang as corresponding authors.
% Dikai Li's real email is pending; the visible placeholder is intentional.
\author{%
Dikai Li \and
Weijia Wang\textsuperscript{*} \and
Yuan-Gen Wang\thanks{Corresponding authors: Weijia Wang and Yuan-Gen Wang.
This work was partially supported by the NSFC 62502108.}%
}

\authorrunning{D.~Li et al.}

\institute{%
School of Artificial Intelligence, Guangzhou University, Guangzhou, China\\
\email{dikaili@e.gzhu.edu.cn, wwj@gzhu.edu.cn, wangyg@gzhu.edu.cn}%
}

\hypersetup{pdfauthor={Dikai Li, Weijia Wang, Yuan-Gen Wang}}

%\inst{1}

\maketitle

This supplement provides the training schedules and reproducibility notes, best-versus-last checkpoint sensitivity, the budget-matched continuation diagnostic, per-step early-exit diagnostics, learned-scale statistics, qualitative fixation examples, and encoder-cost breakdown referenced by the main paper.

\section{Additional Reproducibility Details}
\label{sec:supp_repro}

\Cref{tab:supp_hparams} lists the dataset-specific schedules used for the original main-table checkpoints. Those runs use four focus steps, AdamW, weight decay 0.05, cosine learning rate decay with 5 warmup epochs, label smoothing 0.0, no mixup/cutmix/random erasing, RandAugment, and PPO updates with clip ratio 0.2. The PPO policy standard deviation is scheduled from 0.20 to 0.07 with a cosine decay, except on ImageNet-R, which uses 0.25 to 0.10 and disables the grid-action clipping used elsewhere. In non-exit runs, both the executed scale action and the PPO latent scale coordinate are clamped to the anchor for the first 8 epochs, so the crop field of view stays at the baseline and the scale head is effectively optimized only after warmup; foveation ramps from 0 to 0.5 over epochs 10--20, and detail fusion ramps from 0 to 0.5 over epochs 10--18; early-exit fine-tuning starts from +S+F+D with all three components at full strength. Reported epochs are 1-based throughout. Checkpoints are selected by the public validation/reporting split Acc@1, matching the released training logs. Distributed evaluation follows the released AdaptiveNN-style sampler; when a validation split is not divisible by the number of processes, duplicate padding samples can introduce tiny changes, so noise-level margins in the main paper are not interpreted as accuracy gains. The original accuracy tables are regenerated directly from the training logs by a script included with the code; training scripts and evaluation code will be released.

\begin{table}[tb]
  \caption{Training schedules for the reported runs. BS is effective batch size. EE denotes early-exit fine-tuning from the +S+F+D checkpoint.}
  \label{tab:supp_hparams}
  \centering
  \scriptsize
  \setlength{\tabcolsep}{3.2pt}
  \resizebox{\linewidth}{!}{%
  \begin{tabular}{lrrrrrr}
    \toprule
    Dataset & Non-exit BS & Non-exit epochs & Non-exit LR & EE BS & EE epochs & EE LR \\
    \midrule
    Aircraft & 108 & 80 & 5e-5 & 108 & 20 & 2e-5 \\
    Cars & 108 & 80 & 5e-5 & 108 & 20 & 2e-5 \\
    CUB & 108 & 80 & 5e-5 & 108 & 20 & 2e-5 \\
    Dogs & 108 & 80 & 5e-5 & 108 & 20 & 2e-5 \\
    ImageNet-1K & 144 & 50 & 5e-5 & 144 & 15 & 2e-5 \\
    ImageNet-A & 108 & 60 & 5e-5 & 108 & 20 & 2e-5 \\
    ImageNet-R & 144 & 80 & 1e-4 & 144 & 20 & 5e-5 \\
    Oxford-Pet & 108 & 80 & 5e-5 & 108 & 20 & 2e-5 \\
    \bottomrule
  \end{tabular}}
\end{table}

\begin{table}[tb]
  \caption{External official-codebase rerun versus the same-codebase Orig anchor used for ablations (Acc@1 at the selected epoch). The main paper reports Official as the external fixed-step reference and uses Orig only inside in-codebase ablations; this table records their gap explicitly.}
  \label{tab:supp_orig_official}
  \centering
  \scriptsize
  \setlength{\tabcolsep}{4.2pt}
  \begin{tabular}{lrrr}
    \toprule
    Dataset & Official & Orig & Orig$-$Official \\
    \midrule
    Aircraft & 77.318 & 76.358 & $-0.960$ \\
    Cars & 83.076 & 81.696 & $-1.380$ \\
    CUB & 84.524 & 84.369 & $-0.155$ \\
    Dogs & 91.981 & 92.051 & \gain{0.070} \\
    ImageNet-1K & 82.126 & 82.116 & $-0.010$ \\
    ImageNet-A & 64.815 & 65.873 & \gain{1.058} \\
    ImageNet-R & 85.133 & 85.467 & \gain{0.334} \\
    Oxford-Pet & 96.402 & 96.537 & \gain{0.135} \\
    \bottomrule
  \end{tabular}
\end{table}

\begin{table}[tb]
  \caption{Best-versus-last Acc@1 sensitivity for the main table rows. Each cell is selected-best / final-epoch Acc@1. The middle column selects among component-added non-exit variants, excluding Orig. +S+F ties +S+D on ImageNet-A.}
  \label{tab:supp_best_last}
  \centering
  \scriptsize
  \setlength{\tabcolsep}{4.0pt}
  \resizebox{\linewidth}{!}{%
  \begin{tabular}{lccc}
    \toprule
    Dataset & Orig & Best added non-exit variant & Full+EE \\
    \midrule
    Aircraft & 76.358 / 75.848 & 77.078 / 76.838 (+S+F+D) & 77.558 / 77.258 \\
    Cars & 81.696 / 81.248 & 82.741 / 82.442 (+S+F+D) & 83.711 / 83.648 \\
    CUB & 84.369 / 84.265 & 84.334 / 84.110 (+S+F+D) & 84.800 / 84.455 \\
    Dogs & 92.051 / 91.702 & 92.401 / 91.748 (+S+D) & 92.413 / 92.016 \\
    ImageNet-1K & 82.116 / 81.280 & 82.282 / 80.896 (+S+D) & 82.124 / 81.946 \\
    ImageNet-A & 65.873 / 64.550 & 66.138 / 65.212 (+S+D) & 66.534 / 65.873 \\
    ImageNet-R & 85.467 / 84.367 & 85.900 / 84.867 (+S+D) & 85.633 / 84.900 \\
    Oxford-Pet & 96.537 / 96.266 & 96.456 / 96.131 (+S+D) & 96.591 / 96.591 \\
    \bottomrule
  \end{tabular}}
\end{table}

\Cref{tab:supp_threeseed} reports the three-seed paired results (seeds 0--2) summarized in the main paper; the matched-budget component gain +S+F+D$-$Orig is small but positive on Cars, Aircraft, CUB, and Dogs, and negative-to-zero on Oxford-Pet, which we interpret as near-parity rather than established equivalence.

\begin{table}[tb]
  \caption{Three-seed paired results (seeds 0--2) for the three main arms. Cells are mean Acc@1 $\pm$ standard deviation; the last three columns are paired deltas. These support the seed-0 main table and the near-parity reading of Oxford-Pet.}
  \label{tab:supp_threeseed}
  \centering
  \scriptsize
  \setlength{\tabcolsep}{3.0pt}
  \resizebox{\linewidth}{!}{%
  \begin{tabular}{lcccccc}
    \toprule
    Dataset & Orig & +S+F+D & Full+EE & Full+EE$-$Orig & Full+EE$-$(+S+F+D) & +S+F+D$-$Orig \\
    \midrule
    Aircraft & 76.398$\pm$0.035 & 76.748$\pm$0.330 & 77.308$\pm$0.312 & \gain{0.910}$\pm$0.337 & \gain{0.560}$\pm$0.092 & \gain{0.350}$\pm$0.360 \\
    CUB & 84.210$\pm$0.348 & 84.302$\pm$0.183 & 84.745$\pm$0.187 & \gain{0.535}$\pm$0.165 & \gain{0.443}$\pm$0.020 & \gain{0.092}$\pm$0.176 \\
    Dogs & 92.273$\pm$0.248 & 92.331$\pm$0.107 & 92.514$\pm$0.155 & \gain{0.241}$\pm$0.108 & \gain{0.183}$\pm$0.059 & \gain{0.058}$\pm$0.141 \\
    Oxford-Pet & 96.564$\pm$0.098 & 96.447$\pm$0.031 & 96.537$\pm$0.054 & $-0.027\pm0.098$ & \gain{0.090}$\pm$0.062 & $-0.117\pm0.068$ \\
    Cars & 81.632$\pm$0.457 & 82.731$\pm$0.063 & 84.074$\pm$0.315 & \gain{2.442}$\pm$0.595 & \gain{1.343}$\pm$0.327 & \gain{1.098}$\pm$0.518 \\
    \bottomrule
  \end{tabular}}
\end{table}

\subsection{Paired uncertainty on ImageNet-1K}
\label{sup:paired}

We evaluate fixed Full+EE, +S+F+D, and +S+D checkpoints on all 50,000 validation images with the same prediction-dump protocol (evaluation batch size 48) and align their predictions by sample identity. Their full-step Acc@1 values under this protocol are 82.048\%, 82.188\%, and 82.260\%, respectively. The paired statistics below use these predictions; the main-paper accuracy tables report results from the original evaluation runs.

We obtain unadjusted 95\% percentile confidence intervals from 10,000 paired bootstrap resamples (seed 0). We also apply two-sided exact McNemar tests to the discordant predictions, with Holm correction for the two comparisons with Full+EE. Differences and interval endpoints are in percentage points.

\begin{center}
\small
\begin{tabular}{lrrr}
\toprule
Full+EE minus & Difference & 95\% CI & Exact / Holm $p$ \\
\midrule
+S+F+D & $-0.140$ & $[-0.332,0.056]$ & 0.1611 / 0.1611 \\
+S+D & $-0.212$ & $[-0.404,-0.020]$ & 0.0319 / 0.0638 \\
\bottomrule
\end{tabular}
\end{center}

Neither McNemar comparison reaches the Holm-adjusted 0.05 threshold. The intervals describe image-level sampling uncertainty for the fixed models, whereas variation across training seeds requires separate repeated runs.

\subsection{Training-seed variation on ImageNet-R}
\label{sup:inr_seeds}

We further examine training-seed variation on the fixed ImageNet-R split (split seed 0), using training seeds 0--2. All runs use one GPU with gradient accumulation to an effective batch of 144. This study is separate from the original three-GPU seed-0 runs.

\begin{center}
\small
\begin{tabular}{lc}
\toprule
Variant & Acc@1 (\%, mean $\pm$ sample SD) \\
\midrule
+S+D & $85.167\pm0.379$ \\
+S+F+D & $85.267\pm0.120$ \\
Full+EE & $85.133\pm0.367$ \\
\bottomrule
\end{tabular}
\end{center}

We also compute paired 95\% Student-$t$ intervals across the three seeds, with two degrees of freedom. Differences and interval endpoints are in percentage points.

\begin{center}
\small
\begin{tabular}{lrr}
\toprule
Comparison & Mean difference & 95\% paired interval \\
\midrule
+S+F+D minus +S+D & $+0.100$ & $[-0.622,0.822]$ \\
Full+EE minus +S+F+D & $-0.133$ & $[-1.168,0.901]$ \\
Full+EE minus +S+D & $-0.033$ & $[-1.786,1.719]$ \\
\bottomrule
\end{tabular}
\end{center}

The mean accuracies are close, but the three-seed intervals remain wide. This study characterizes the observed seed variation without establishing equivalent performance.

\section{Budget-Matched Continuation Diagnostic}
\label{sec:supp_budget}

\Cref{tab:budget_diag} reports the budget-matched continuation summarized in the main paper: a 30-epoch single-GPU continuation of Orig, +S+F+D, and Full+EE with matched effective batch and learning-rate schedule. Extra training of the fixed-step Orig model is itself a strong accuracy baseline, so once the budget is matched the proposed components yield dataset-dependent gains rather than a uniform improvement. This controls only the accuracy axis and does not give the fixed-step model the learned per-fixation scale or the confidence-conditioned operating points of the early-exit frontier.

\begin{table}[tb]
  \caption{Budget-matched continuation diagnostic: 30-epoch single-GPU continuation with matched effective batch and learning-rate schedule (Acc@1/Acc@5 at the selected epoch). $\Delta$Orig+30 and $\Delta$+S+F+D+30 are Full+EE+30 minus the corresponding continued baseline. Scale-cost+30 is a diagnostic +S+F+D continuation with an auxiliary scale-cost penalty; it is not used by any reported main-paper run and was run only for Aircraft, CUB, and ImageNet-A. This diagnostic was run only for Aircraft, CUB, and ImageNet-A.}
  \label{tab:budget_diag}
  \centering
  \scriptsize
  \setlength{\tabcolsep}{3.0pt}
  \resizebox{\linewidth}{!}{%
  \begin{tabular}{lcccccc}
    \toprule
    Dataset & Orig+30 & +S+F+D+30 & Full+EE+30 & Scale-cost+30 & $\Delta$Orig+30 & $\Delta$+S+F+D+30 \\
    \midrule
    Aircraft & 79.718/96.910 @24 & 80.168/97.120 @30 & \best{80.378}/97.000 @28 & 79.748/97.030 @27 & \gain{0.660} & \gain{0.210} \\
    Cars & 86.121/98.222 @27 & 86.668/98.408 @25 & \best{86.880}/98.346 @24 & -- & \gain{0.759} & \gain{0.211} \\
    CUB & 85.554/97.428 @30 & 85.623/97.204 @30 & \best{85.692}/97.273 @24 & 85.433/97.290 @29 & \gain{0.138} & \gain{0.069} \\
    Dogs & 92.296/99.510 @4 & 92.226/99.510 @5 & \best{92.378}/99.499 @3 & -- & \gain{0.082} & \gain{0.152} \\
    ImageNet-A & \best{67.328}/82.407 @23 & 66.402/82.804 @18 & 66.005/82.011 @15 & 66.270/81.746 @29 & $-1.323$ & $-0.397$ \\
    Oxford-Pet & \best{96.645}/99.919 @22 & 96.429/99.946 @20 & 96.456/99.756 @14 & -- & $-0.189$ & \gain{0.027} \\
    \bottomrule
  \end{tabular}}
\end{table}

\section{Early-Exit Diagnostics}
\label{sec:supp_earlyexit}

\subsection{Early exit initialized from +S+D}
\label{sup:no_foveation}

We fine-tune early-exit models from +S+D and +S+F+D on ImageNet-R for 20 epochs at learning rate $5\times10^{-5}$. The effective training batches are 48 and 144, respectively. The models therefore differ in both their starting configurations and training batches.

We fit stopping thresholds by five-fold cross-fitting on the same 3,000-image evaluation split and apply the rules to full-rollout predictions. The table reports Acc@1 of the +S+D-initialized model minus that of the +S+F+D-initialized model, in percentage points. The largest difference in simulated average exit steps between configurations is 0.00534.

\begin{center}
\small
\begin{tabular}{lrrrr}
\toprule
Stopping rule / target steps & 1.5 & 2.0 & 2.5 & 3.0 \\
\midrule
Max-softmax & $+0.167$ & $+0.067$ & $0.000$ & $-0.067$ \\
Learned stop head & $-0.167$ & $-0.300$ & $-0.167$ & $-0.433$ \\
\bottomrule
\end{tabular}
\end{center}

The +S+D-initialized model does not consistently improve accuracy across stopping rules and budgets. The comparison evaluates an alternative early-exit configuration, with training-batch differences retained as a source of confounding.

\Cref{tab:supp_steps} reports per-step Acc@1 for the full early-exit checkpoint evaluated without early stopping. Accuracy generally increases across focus steps, supporting the use of confidence-conditioned stopping as an inference-time operating point.

\begin{table}[tb]
  \caption{Per-step Acc@1 for the full early-exit checkpoint without early stopping.}
  \label{tab:supp_steps}
  \centering
  \scriptsize
  \setlength{\tabcolsep}{5.0pt}
  \begin{tabular}{lrrrr}
    \toprule
    Dataset & Step 1 & Step 2 & Step 3 & Step 4 \\
    \midrule
    Aircraft & 71.347 & 74.377 & 76.508 & 77.558 \\
    Cars & 76.200 & 80.291 & 82.753 & 83.711 \\
    CUB & 82.678 & 83.834 & 84.438 & 84.800 \\
    Dogs & 90.688 & 91.713 & 92.040 & 92.413 \\
    ImageNet-1K & 79.704 & 81.106 & 81.810 & 82.040 \\
    ImageNet-A & 56.746 & 61.772 & 63.757 & 66.534 \\
    ImageNet-R & 82.033 & 83.700 & 84.533 & 85.633 \\
    Oxford-Pet & 95.942 & 95.969 & 96.402 & 96.591 \\
    \bottomrule
  \end{tabular}
\end{table}

\Cref{tab:supp_scale} gives the mean learned scale over the four focus steps for the same checkpoints. Aircraft, CUB, Dogs, and ImageNet-A mostly stay near the anchor scale, while Cars, ImageNet-1K, ImageNet-R, and Oxford-Pet use larger average fields of view.

\begin{table}[tb]
  \caption{Mean learned scale per focus step for the full early-exit checkpoint without early stopping. The last column is the standard deviation of the learned scale across validation images, averaged over the four steps.}
  \label{tab:supp_scale}
  \centering
  \scriptsize
  \setlength{\tabcolsep}{4.0pt}
  \begin{tabular}{lrrrrr}
    \toprule
    Dataset & Step 1 & Step 2 & Step 3 & Step 4 & Avg. std \\
    \midrule
    Aircraft & 1.006 & 1.015 & 1.011 & 1.008 & 0.007 \\
    Cars & 1.110 & 1.167 & 1.167 & 1.215 & 0.077 \\
    CUB & 1.006 & 1.010 & 1.008 & 1.012 & 0.008 \\
    Dogs & 1.002 & 1.003 & 1.002 & 1.002 & 0.002 \\
    ImageNet-1K & 1.078 & 1.162 & 1.175 & 1.106 & 0.055 \\
    ImageNet-A & 1.004 & 1.003 & 1.004 & 1.003 & 0.003 \\
    ImageNet-R & 1.150 & 1.155 & 1.194 & 1.175 & 0.059 \\
    Oxford-Pet & 1.055 & 1.076 & 1.051 & 1.070 & 0.058 \\
    \bottomrule
  \end{tabular}
\end{table}

\begin{figure}[tb]
  \centering
  \includegraphics[width=\linewidth]{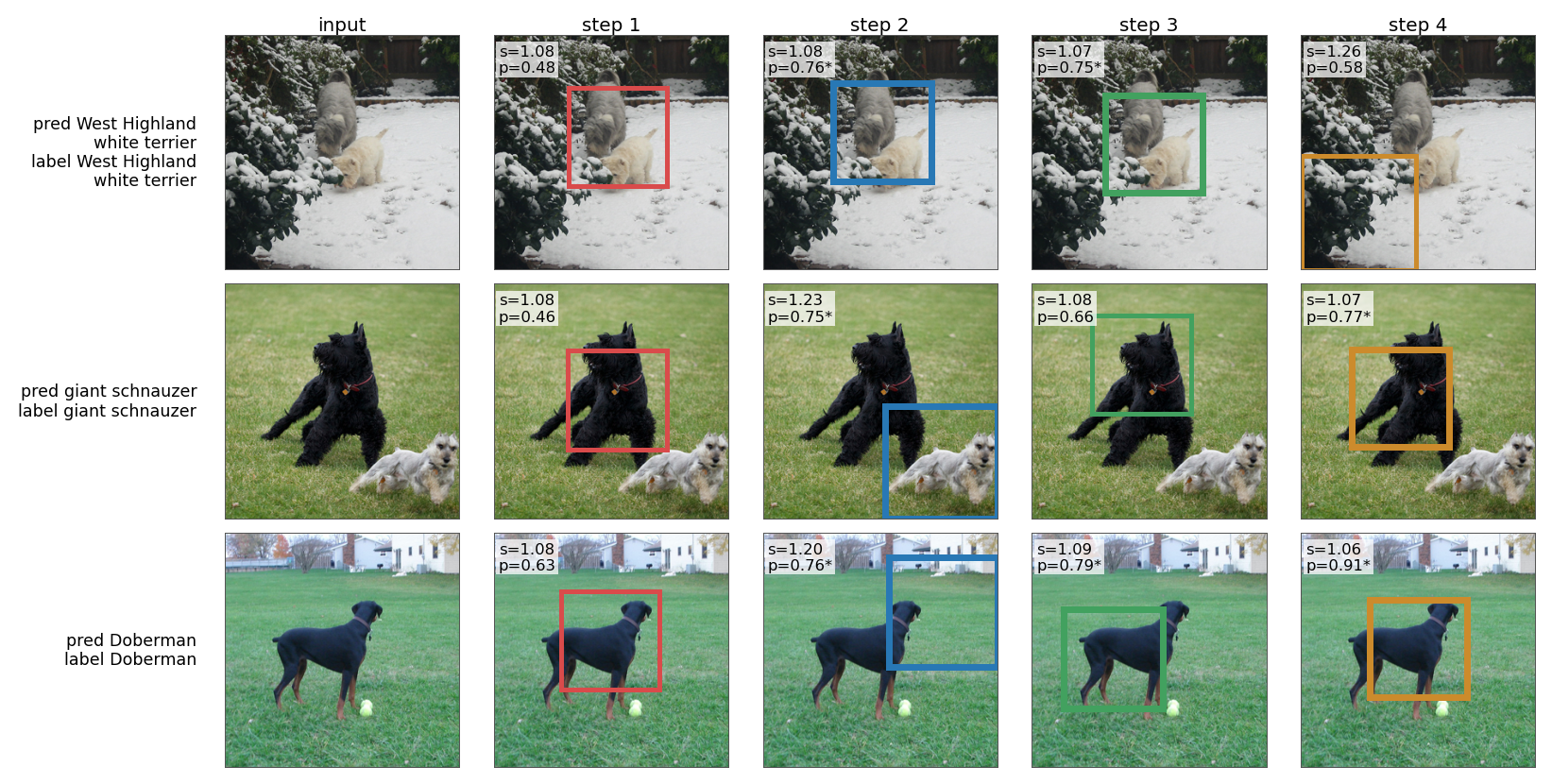}
  \caption{Qualitative fixation trajectories on ImageNet validation examples at $\tau=0.70$. Each row shows the input and four predicted focus boxes from the Full+EE checkpoint. The overlay reports learned scale $s$ and stop probability $p=\sigma(q_t)$; asterisks identify the displayed stop-probability annotations.}
  \label{fig:supp_qualitative_fixations}
\end{figure}

\begin{figure}[tb]
  \centering
  \includegraphics[width=0.96\linewidth]{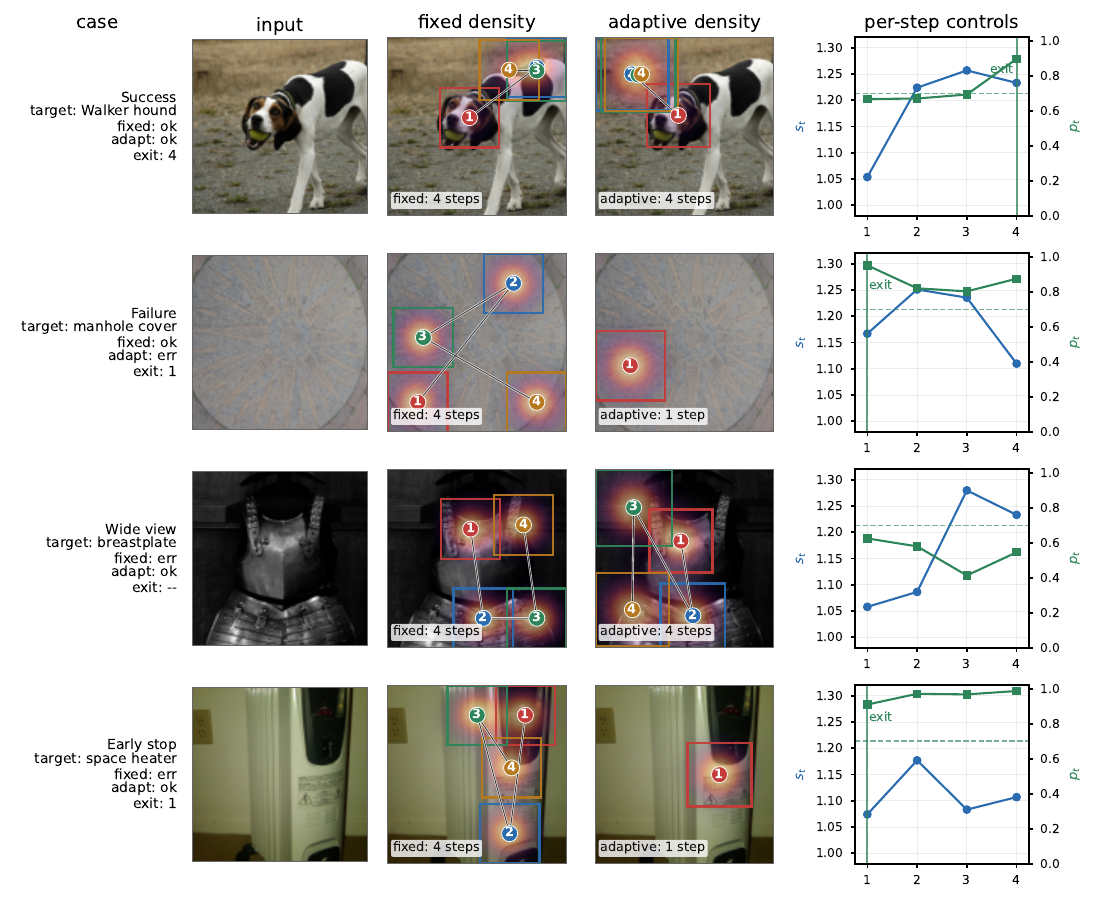}
  \caption{Additional ImageNet-1K fixation-density contrasts, including the adaptive failure case. Fixed density accumulates four fixed-scale crops; adaptive density accumulates only the crops executed at $\tau=0.70$. Heat intensity remains a per-example fixation-density map rather than a dataset-level fixation distribution.}
  \label{fig:supp_fixation_density_full}
\end{figure}

\begin{figure}[tb]
  \centering
  \includegraphics[width=0.9\linewidth]{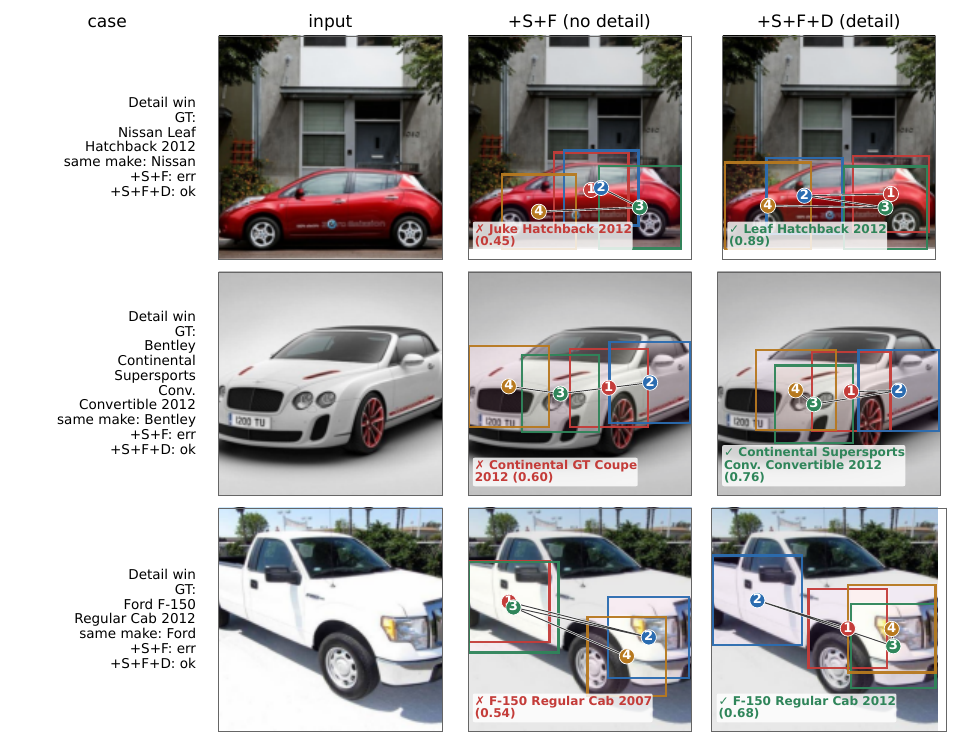}
  \caption{Stanford Cars validation cases where detail fusion flips the prediction. Each row shows the input and the executed four-step fixations of the no-detail (+S+F) and detail (+S+F+D) ablation checkpoints; numbered boxes mark fixation order, heat shading marks accumulated crop coverage, and each label reports the prediction with its confidence. All three cases are same-make confusions corrected by the detail-enabled checkpoint.}
  \label{fig:supp_detail_contrast}
\end{figure}

\Cref{fig:supp_detail_contrast} illustrates the detail-token contribution on Stanford Cars, the dataset with the most consistent detail gains. The comparison uses +S+F and +S+F+D; both enable scale and foveation and execute four focus steps, with detail enabled only in the latter. A deterministic re-evaluation of the 8,041-image validation split finds 379 examples where detail fusion corrects a no-detail error against 290 reverse cases. The shown examples are same-make confusions (Leaf vs.\ Juke, Continental trim variants, F-150 model years) whose discriminative evidence is fine-grained. An analogous comparison of foveation configurations (+S+D versus +S+F+D, detail branch fixed) flips 328 validation examples right against 290 reverse cases and supplies the foveation row of the main-paper qualitative figure.

\begin{figure}[tb]
  \centering
  \includegraphics[width=0.86\linewidth]{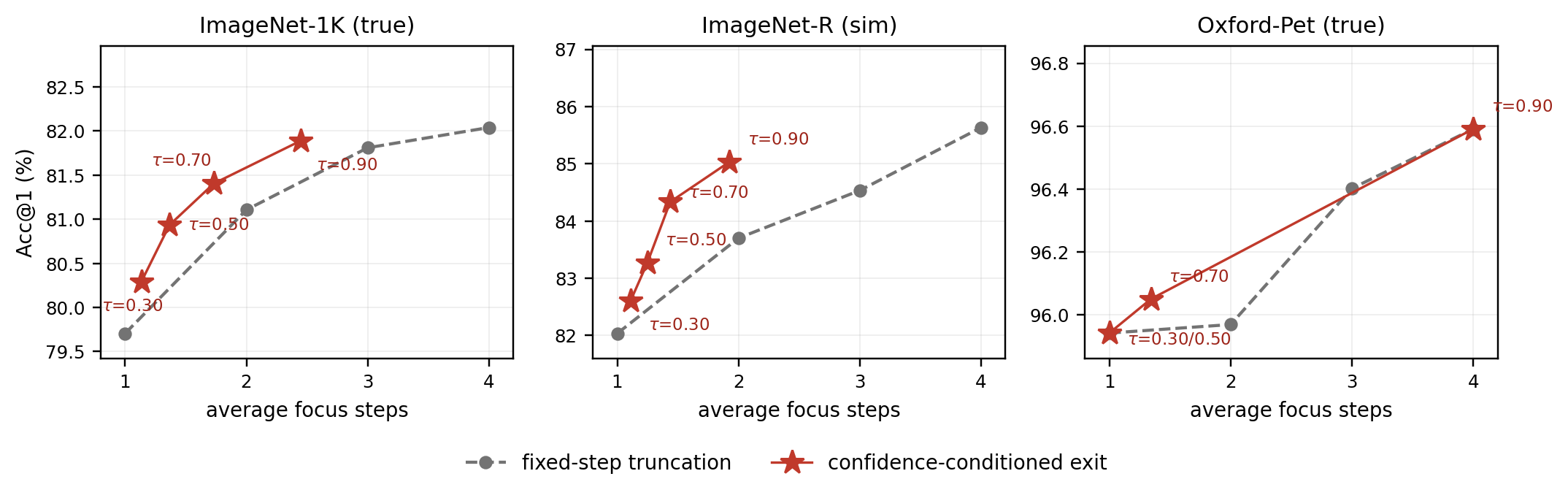}
  \caption{Accuracy versus average focus steps for the Full+EE checkpoint. Dashed: truncating every sequence at a fixed step (1--4). Red markers: confidence-conditioned exit for $\tau\in\{0.30,0.50,0.70,0.90\}$, using realized sequential-step measurements where available and otherwise the full-forward stop-rule simulation. Sequential-step points measure realized exits; simulated points analyze the stop rule without claiming realized savings.}
  \label{fig:supp_frontier_full}
\end{figure}

\Cref{fig:supp_frontier_full} reproduces the full accuracy-compute diagnostic: confidence-conditioned exits against fixed-step truncation of the same Full+EE checkpoint. The main paper isolates the realized sequential panels; here, ImageNet-R is included as a simulation-only stop-rule diagnostic.

\begin{figure}[tb]
  \centering
  \includegraphics[width=0.46\linewidth]{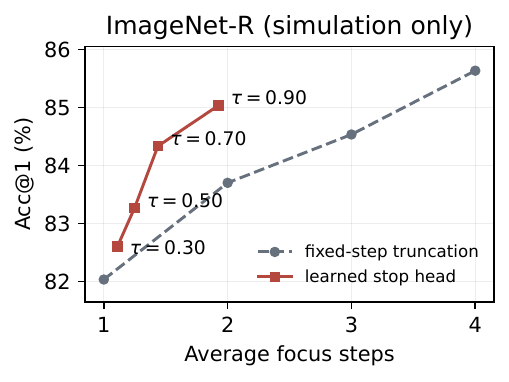}
  \caption{Simulation-only ImageNet-R stop-rule diagnostic from full four-step rollouts. This panel helps separate the shape of the learned stopping criterion from realized sequential FLOP savings, whose estimates use physically executed step counts only for Seq. rows in the main table.}
  \label{fig:supp_frontier_sim}
\end{figure}

\FloatBarrier
\section{Computational Cost}
\label{sec:supp_flops}

The cost breakdown uses a conservative encoder-pass estimate. Each glance, focus, or detail pass contributes 1.10 GFLOPs under the DeiT-S multiply-add convention used for this breakdown. For Full+EE, the estimate is $1.10+2.20\bar{K}$ GFLOPs, where $\bar{K}$ is the average number of selected focus steps. It counts one detail pass at every selected step, including the first, although the implementation enables detail only from the second focus step onward. Four-step estimates are therefore 5.50 GFLOPs for Orig and 9.90 GFLOPs for Full+EE under the conservative convention. Crop sampling, policy and classification heads, feature recovery, and other non-encoder operations are outside this estimate. The Seq. rows use physically executed step counts; Sim. rows give analytic costs for stops selected from full four-step rollouts and do not measure realized savings.

\begin{table}[tb]
  \caption{Conservative encoder-pass cost estimate (GFLOPs, multiply-adds; non-encoder operations excluded) of the Full+EE checkpoint at each early-exit threshold, computed as glance + average steps $\times$ (focus + detail). Simulated rows are analytic stop-rule costs rather than realized sequential savings. The original model at four fixed steps costs 5.50 GFLOPs under the same proxy.}
  \label{tab:supp_flops}
  \centering
  \scriptsize
  \setlength{\tabcolsep}{3.8pt}
  \begin{tabular}{llrrrrr}
    \toprule
    & & \multicolumn{2}{c}{$\tau=0.70$} & \multicolumn{2}{c}{$\tau=0.90$} & Full \\
    \cmidrule(lr){3-4} \cmidrule(lr){5-6}
    Dataset & Protocol & Avg. step & GFLOPs & Avg. step & GFLOPs & GFLOPs \\
    \midrule
    Aircraft & Sim. & 3.67 & 9.17 & 4.00 & 9.90 & 9.90 \\
    Cars & Sim. & 3.99 & 9.87 & 4.00 & 9.90 & 9.90 \\
    CUB & Sim. & 3.71 & 9.26 & 4.00 & 9.90 & 9.90 \\
    Dogs & Sim. & 3.99 & 9.88 & 4.00 & 9.90 & 9.90 \\
    ImageNet-1K & Seq. steps & 1.73 & 4.91 & 2.45 & 6.48 & 9.90 \\
    ImageNet-A & Sim. & 3.96 & 9.81 & 4.00 & 9.90 & 9.90 \\
    ImageNet-R & Sim. & 1.44 & 4.26 & 1.93 & 5.33 & 9.90 \\
    Oxford-Pet & Seq. steps & 1.34 & 4.05 & 4.00 & 9.89 & 9.90 \\
    \bottomrule
  \end{tabular}
\end{table}

\subsection{Model-forward runtime and memory}
\label{sup:runtime}

We profile ImageNet-1K inference on an NVIDIA RTX 4090 with PyTorch 2.5.1+cu121 and CUDA 12.1. CUDA-event timing covers model-forward execution and excludes disk I/O, preprocessing, and host-to-device transfer. Each policy is measured after 30 warmup batches, followed by 100 timed batches.

For adaptive inference, we use classifier max-softmax as the stopping score. At each target step budget, five-fold threshold fitting produces five thresholds, whose mean is used as one fixed threshold for profiling. At the two-step target, this threshold is 0.914015, and the timed batch-1 images execute 2.04 focus steps on average. Timing uses this single threshold and a profiling subset; accuracy evaluation with fold-specific thresholds is a separate protocol.

\begin{center}
\small
\begin{tabular}{lrrr}
\toprule
Batch-1 policy & Mean (ms) & P90 (ms) & \shortstack{Peak allocated\\(MiB)} \\
\midrule
Orig, fixed 4 & 52.23 & 57.61 & 543.46 \\
+S+D, fixed 4 & 71.88 & 79.20 & 547.06 \\
Full+EE, fixed 2 & 37.63 & 41.56 & 515.91 \\
Full+EE, adaptive target 2 & 38.40 & 74.55 & 547.64 \\
Full+EE, fixed 4 & 71.33 & 75.53 & 547.64 \\
\bottomrule
\end{tabular}
\end{center}

Adaptive inference reduces mean batch-1 latency relative to fixed four-step Full+EE inference, while P90 changes little. Fixed two-step inference has similar mean latency and a lower P90.

\begin{center}
\small
\begin{tabular}{lrrr}
\toprule
Policy & \shortstack{Batch 32\\(images/s)} & \shortstack{Batch 48\\(images/s)} & \shortstack{Batch 48 peak\\(MiB)} \\
\midrule
Orig, fixed 4 & 562.1 & 855.1 & 718.85 \\
+S+D, fixed 4 & 411.4 & 616.0 & 732.14 \\
Full+EE, fixed 2 & 756.6 & 1115.9 & 691.19 \\
Full+EE, adaptive target 2 & 409.0 & 620.3 & 701.59 \\
Full+EE, fixed 4 & 423.5 & 613.1 & 734.37 \\
\bottomrule
\end{tabular}
\end{center}

The batch-32 and batch-48 measurements cover 3,200 and 4,800 timed images, respectively. Adaptive execution has no consistent throughput advantage over fixed four-step Full+EE inference in this implementation. The Orig and +S+D rows provide additional runtime controls under the same profiler.

\subsection{Recorded training cost}
\label{sup:training_cost}

In the recorded three-GPU ImageNet-1K +S+D run, training takes approximately 1.4 wall-clock hours per epoch, corresponding to 4.2 GPU-hours. The 50-epoch run therefore costs approximately 210 GPU-hours. The logged early-exit fine-tuning epochs take approximately 1.7 wall-clock hours each on three GPUs, or 5.1 GPU-hours per epoch. The early-exit figure is a per-epoch estimate from the available log segment.